%% file: root.tex
\documentclass[letterpaper, 10 pt, conference]{ieeetran}  

\IEEEoverridecommandlockouts                              

\input{includes/packages}

\input{includes/macros}
\newcommand{\MethodName}{\textbf{\texttt{RAVEN}}}

\title{\LARGE \bf RAVEN: Resilient Aerial Navigation via\\ Open-Set Semantic Memory and Behavior Adaptation}

\author{Seungchan Kim$^{1}$, Omar Alama$^{1}$, Dmytro Kurdydyk$^{2}$, John Keller$^{1}$, \\ Nikhil Keetha$^{1}$, Wenshan Wang$^{1}$, Yonatan Bisk$^{1}$, Sebastian Scherer$^{1}$
\thanks{$^{1}$ Authors are with Carnegie Mellon University, Pittsburgh, PA, USA. \tt{\{seungch2, oalama, jkeller2, nkeetha, wenshanw, ybisk, basti\}@andrew.cmu.edu}}
\thanks{$^{2}$ Author is with Davidson College, Davidson, NC, USA. \tt{dmkurdydyk@davidson.edu}}
}

\begin{document}
\thispagestyle{empty}
\pagestyle{empty}

\makeatletter
\let\@oldmaketitle\@maketitle
\renewcommand{\@maketitle}{\@oldmaketitle
\centering
\captionsetup{type=figure, singlelinecheck=false,font=small}
\begin{tabular}{cccc}
\includegraphics[width=0.97\textwidth,trim=5mm 17mm 10mm 0mm]{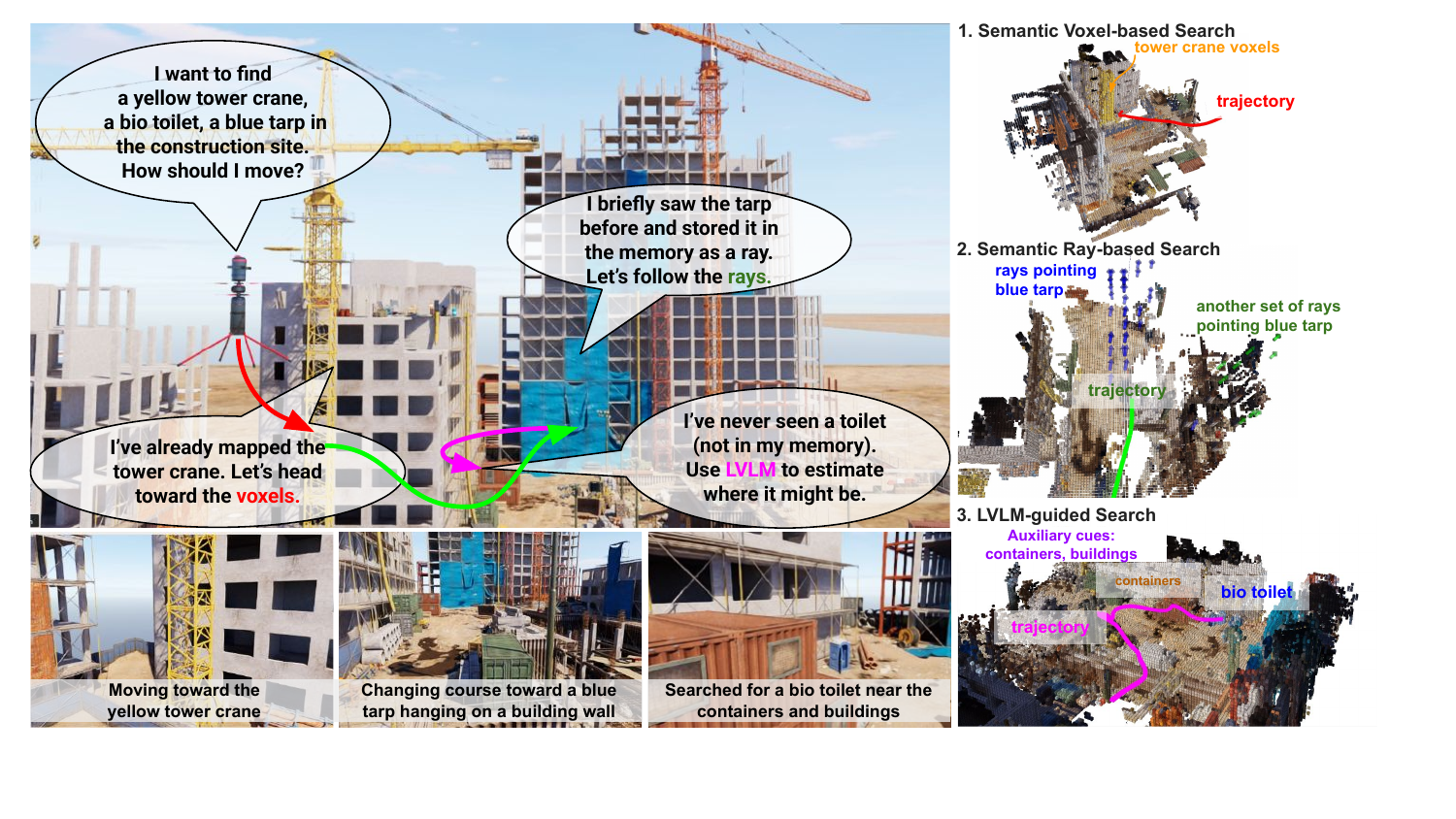}
\end{tabular}
\caption{We present \MethodName, a 3D open-set memory-based behavior tree framework for aerial semantic outdoor navigation. \MethodName~not only navigates reliably toward detected targets, but also performs long-range reasoning to plan toward distant cues and continues informed search even when direct visual evidence is limited. It further supports multi-class object search and on-the-fly task switching within a mission.} 
\label{fig:intro}
\setcounter{figure}{1}
}


\makeatother

\maketitle

\begin{abstract}
Aerial outdoor semantic navigation requires robots to explore large, unstructured environments to locate target objects. Recent advances in semantic navigation have demonstrated open-set object-goal navigation in indoor settings, but these methods remain limited by constrained spatial ranges and structured layouts, making them unsuitable for long-range outdoor search. While outdoor semantic navigation approaches exist, they either rely on reactive policies based on current observations, which tend to produce short-sighted behaviors, or precompute scene graphs offline for navigation, limiting adaptability to online deployment. We present \MethodName, a 3D memory-based, behavior tree framework for aerial semantic navigation in unstructured outdoor environments. It (1) uses a spatially consistent semantic voxel-ray map as persistent memory, enabling long-horizon planning and avoiding purely reactive behaviors, (2) combines short-range voxel search and long-range ray search to scale to large environments, (3) leverages a large vision-language model to suggest auxiliary cues, mitigating sparsity of outdoor targets. These components are coordinated by a behavior tree, which adaptively switches behaviors for robust operation. We evaluate \MethodName~in 10 photorealistic outdoor simulation environments over 100 semantic tasks, encompassing single-object search, multi-class, multi-instance navigation and sequential task changes. Results show \MethodName~outperforms baselines by $85.25\%$ in simulation and demonstrate its real-world applicability through deployment on an aerial robot in outdoor field tests.\\Website: \textcolor{blue}{ \href{https://raven-semantic.github.io/}{https://raven-semantic.github.io/}}
\end{abstract}


\section{INTRODUCTION}
Aerial robots are increasingly employed for autonomous exploration and search in outdoor environments. Prior research has largely focused on geometry-driven strategies, including frontier-based exploration \cite{zhou2021fuel} and volumetric information gain maximization \cite{bircher2016receding}. While these methods efficiently cover the spaces, they rely on geometric representations--such as occupancy grids or volumetric maps--and cannot perform semantic reasoning, for example, locating a yellow car in urban areas or identifying machinery scattered across a construction site.

Semantic navigation has been extensively studied in indoor environments~\cite{chaplot2020object, ramakrishnan2022poni, chang2023goat}, where robots are tasked with reaching object goals or following language instructions. With the advent of foundation models, recent approaches have leveraged these models for open-set object navigation \cite{yokoyama2024vlfm, dorbala2022clipnav, gadre2023cows}. However, such methods remain largely confined to bounded 2D indoor settings \cite{chaplot2020object,ramakrishnan2022poni,yokoyama2024vlfm, chang2023goat,dorbala2022clipnav, gadre2023cows}, where the spatial horizon is limited, the robot’s range of possible movements is constrained, and the structured environment provides cues for locating objects.

By contrast, outdoor semantic navigation poses two major challenges compared to indoor settings. First, the much larger spatial extent of outdoor scenes requires long-range search strategies; second, the sparse distribution of target objects, combined with the absence of a structured hierarchy (e.g., building $\rightarrow$ floor $\rightarrow$ room $\rightarrow$ object) demands careful strategic planning. Some prior methods~\cite{liu2023aerialvln, gao2025openfly,xiao2025uav} pursue map-free navigation with reactive policies that leverage short-term observation histories, which often generate short-sighted behaviors. Other works attempt semantic navigation in outdoor scenes using graph representations; however, they construct these graphs offline, which limits their online applicability~\cite{xie2024embodied,strader2025language}.

We propose \MethodName, a \textbf{R}esilient \textbf{A}erial \textbf{V}oxel-Ray Memory \textbf{E}mpowered \textbf{N}avigation, a novel framework for outdoor semantic search. (1) \MethodName~builds upon recent advances in open-set semantic voxel and ray representations~\cite{alama2025rayfronts}, leveraging them as \textit{persistent internal memory}. This open-set voxel-ray memory not only avoids purely reactive behaviors and enables long-horizon planning, but also supports multi-class search and accommodates dynamic task switching, owing to its task-agnostic nature. (2) Beyond voxel-based search for close and reliable cues, \MethodName~performs ray-based search to register and plan over long-range distance cues, thereby addressing the challenge of long-range reasoning in outdoor settings. (3) To overcome the semantic sparsity of outdoor scenes, it invokes a large vision language model (LVLM)-guided search when needed to incorporate auxiliary cues. These search behaviors are integrated through a behavior tree, which allows the system to adaptively select behaviors depending on situations and to maintain resiliency by switching strategies when one fails. We validate \MethodName~in extensive simulation environments, and demonstrate its promising performance through real-robot deployment testing.

In summary, our contributions are as follows:
\begin{itemize}
    \item We present a new aerial semantic navigation framework that leverages an open-set voxel–ray memory as an internal representation of the world and employs a behavior tree to robustly adapt across multiple search behaviors.  
    \item We address the challenges of long-range reasoning and sparse semantic cues via ray-based search and LVLM guidance within the behavior tree. Our method further supports multi-class search and on-the-fly task switching, an underexplored aspect of semantic navigation. 
    \item We validate the framework by demonstrating superior performance over existing baselines across diverse tasks in simulation and further showcase its real-world feasibility through deployment on a physical robot.

\end{itemize}

\section{RELATED WORKS}
Mobile robot autonomy spans diverse tasks, from exploration to navigation, and building on these foundations, recent works have explored the semantic navigation paradigm \cite{ chaplot2020object, ramakrishnan2022poni,chang2023goat}, where robots understand and reason about semantic information in the environment, such as object types, language, and context, to guide their navigation.

With the rise of Vision Foundation Models (VFMs) \cite{radford2021learning} and LVLMs \cite{zhu2025internvl3}, semantic navigation has seen tremendous growth \cite{zhou2024navgpt,ren2024explore}. Tasks in semantic navigation include vision-language navigation (VLN) \cite{anderson2018vision, zhou2024navgpt, cheng2024navila}, where a robot executes a vision-grounded language instruction to follow paths; object-goal navigation (OGN) \cite{chaplot2020object, yokoyama2024vlfm, wani2020multion}, where a robot locates and navigates toward objects based on their categories or descriptions; embodied question answering (EQA) \cite{ren2024explore}, where a robot actively explores to answer questions about the environment. 

In this work, we focus on object-goal navigation. While prior studies have primarily focused on indoor, ground robots, and single-object scenarios, we aim to explore OGN in outdoor settings, with aerial robots, and multi-class, multi-object scenarios--an important yet widely underexplored domain. 

\subsection{Classical Exploration and Search}
Early work on robotic exploration began with frontier-based methods using 2D occupancy grids \cite{yamauchi1997frontier}. Extensions to 3D aerial scenarios include frontier-based fast exploration \cite{zhou2021fuel} and sampling-based approaches driven by information gain \cite{bircher2016receding}. While these methods have shown strong performance in real-world deployments, their primary focus is on mapping and covering unknown areas, often neglecting semantic information. Recent efforts \cite{ best2022resilient} employ aerial robots to explore regions and attempt to detect target objects. However, they treat object discovery merely as a byproduct of map coverage rather than performing object-goal-driven planning. In contrast, \MethodName~leverages open-set semantic voxel-ray memory and LVLM to guide planning, while retaining the robustness of traditional methods when semantic guidance is insufficient.

\subsection{Map-free Semantic Navigation}
Building on the success of vision language models (VLMs) and multi-modal LLMs, recent works explored zero-shot semantic navigation that uses semantic priors to interpret observations and directly generate control commands. Early approaches relied on contrastive VLMs such as CLIP \cite{radford2021learning} to sequentially guide the robot \cite{dorbala2022clipnav}. Later works leverage LLMs' reasoning capabilities \cite{weerakoon2025behav} and emphasize incorporating memory to go beyond reactive planning, e.g., by buffering past FPV frames \cite{cheng2024navila, gao2025openfly} or maintaining textual summaries of past observations and actions \cite{zhou2024navgpt}. This map-free paradigm has also been applied to VLN \cite{liu2023aerialvln, gao2025openfly} and OGN \cite{xiao2025uav} in aerial robotics.

However, the FPV-based map-free paradigm typically produces discrete control commands, bypassing motion continuity and showing limitations in collision avoidance. Moreover, relying on latest frames or heuristically selected history suffers from information loss \cite{zhou2024navgpt}, limiting applicability to long-horizon planning. In contrast, our approach leverages a 3D spatially grounded open-set semantic memory that efficiently aggregates in-range and out-of-range observations into a persistent internal representation, enabling long-horizon planning.

\begin{figure*}[ht!]
    \centering
\includegraphics[width=\linewidth]{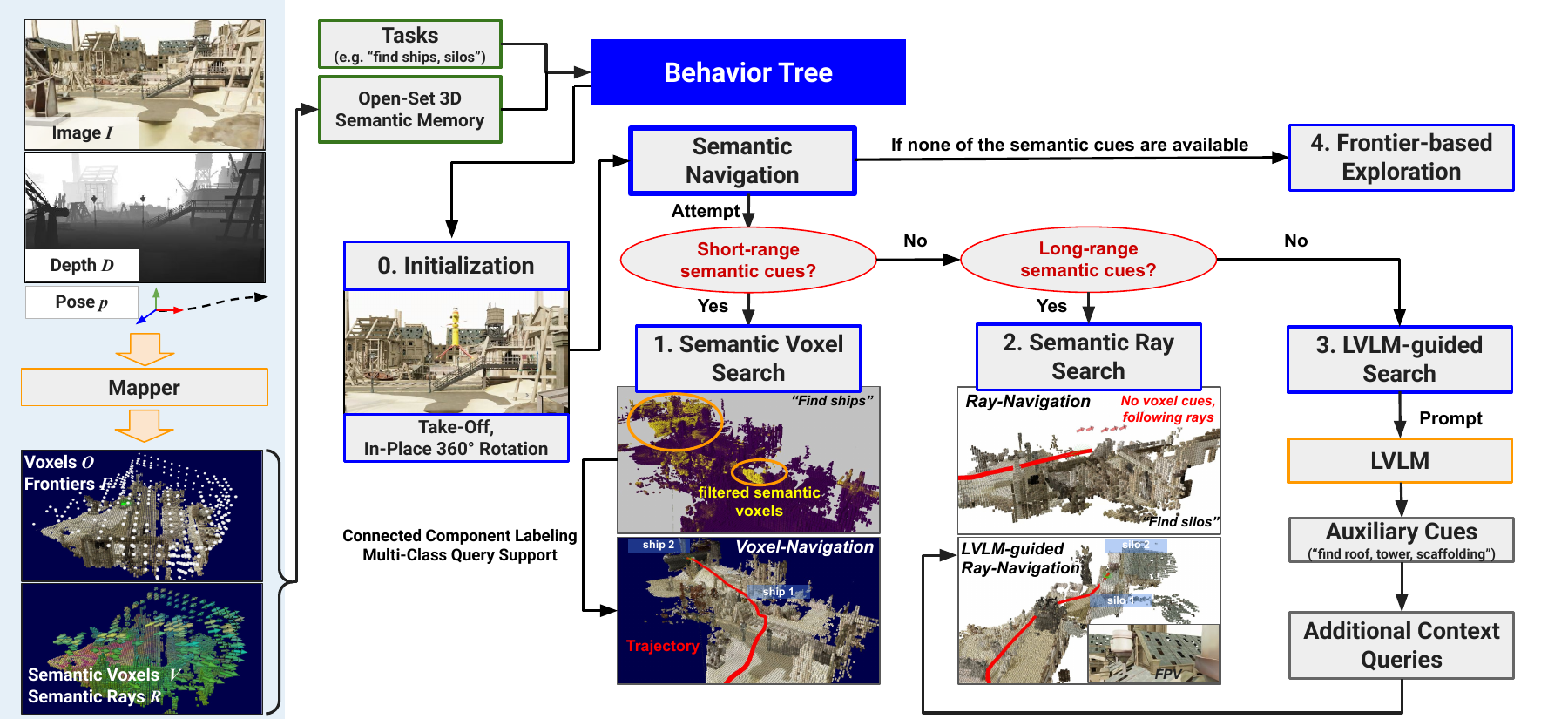} 
\captionsetup{font=small}
\caption{Overview of \MethodName. From image, depth, and pose inputs, the mapper builds an open-set 3D semantic voxel-ray map that serves as persistent memory. A behavior tree adapts the robot's actions: it performs semantic voxel-based search when reliable cues exist within depth range, switches to semantic ray-based search when only long-range directional hints are available, invokes an LVLM to suggest auxiliary objects when no target is visible, and defaults to frontier-based exploration if all strategies fail.}
    \label{Method}
\end{figure*}

\subsection{Map-based Semantic Navigation}
To address limitations of LLMs' spatial reasoning \cite{spatialvlm} and support longer-term observation storage, many works adopt spatial mapping approaches. In these methods, semantic information from observations is encoded and aggregated into either: (1) dense representations, such as 2D bird's eye view (BEV) grids--often combined with semantically scored frontier maps \cite{yokoyama2024vlfm, chaplot2020object, ramakrishnan2022poni, chang2023goat}--or 3D voxel grids \cite{conceptfusion, alama2025rayfronts}; or (2) graph-based representations, including scene graphs that associate semantics with scene entities \cite{gu2024conceptgraphs,hovsg, rana2023sayplan}.

Graph-based approaches have a lower memory footprint but are often computationally expensive to construct, with many generated offline before use \cite{conceptfusion,gu2024conceptgraphs,hovsg,rana2023sayplan, xie2024embodied, strader2025language}. Furthermore, their heuristic-driven graph structures typically assume regular indoor layouts, making them difficult to generalize to outdoor environments \cite{deng2024opengraph}. Similarly, 2D BEV-based approaches \cite{yokoyama2024vlfm, chaplot2020object, ramakrishnan2022poni, chang2023goat, gadre2023cows}, while capable of online guidance, are largely applied to indoor ground-robot scenarios, assuming constrained spatial ranges and hierarchical indoor structures.  

Recently, RayFronts \cite{alama2025rayfronts} introduced a real-time, online mapping framework that generates a dense semantic representation using a single forward pass. It encodes in-range observations as voxels and out-of-range observations as rays, enabling object localization given a \textbf{fixed trajectory} and \textbf{explicit queries}. We leverage this representation as a memory to \textbf{actively guide} an aerial robot toward goal objects and \textbf{generate reasoning-based queries} via an LVLM, enabling the system not only to navigate toward targets, but also to strategically identify auxiliary cues to support efficient search.

\section{METHOD}
The core of \MethodName~is a behavior tree that adapts search behaviors, leveraging a 3D open-set semantic voxel-ray representation as a memory. We first provide a brief overview of the voxel-ray mapping, and then focus on behavior modules in detail. The overall pipeline is illustrated in Figure~\ref{Method}.

\subsection{Preliminary: Open-Set Semantic Voxel-Ray Mapping}
We build upon the encoder and the open-set 3D voxel and ray mapping pipeline from RayFronts~\cite{alama2025rayfronts}. Specifically, at each timestep $t$ the robot receives an RGB image $I_t$ and depth measurement $D_t$. The encoder $E$ processes the image to produce vision-language aligned features $\mathbf{f}_t = E(I_t)$. From the depth sensor, we construct a set of 3D occupancy voxels $O_t$ and detect frontier regions $\mathcal{F}_t$. The image features are projected onto the voxel grid, resulting in semantic voxels:
\begin{equation}
    V_t = \textsc{Project}(\mathbf{f}_t, O_t).
\end{equation}
In parallel, we cast outward rays from $\mathcal{F}_t$, and similarly project the features $\mathbf{f}_t$ onto them, yielding semantic rays:
\begin{equation}
    R_t = \textsc{RayProject}(\mathbf{f}_t, \mathcal{F}_t).
\end{equation}
Rays serve to encode semantic information beyond the depth coverage, complementing $V_t$. Unlike a frontier $f \in \mathcal{F}_t$, which is represented as a single 3D point, each ray $r_i \in R_t$ has its own 3D origin and \textit{directional} vector. Multiple rays may share the same frontier as their origin, with each ray pointing a different direction. These directions are computed via image projection, preserving accurate orientation. Rays also provide a much sparser representation than the full set of image observations, enabling efficient encoding of distant semantic cues.

\subsection{\MethodName: Behavior Tree for Resilient Adaptation}
\label{global-planner}
This subsection presents our main contribution: a set of modular behaviors for 3D aerial outdoor semantic navigation. We describe each module, and conclude by explaining how they are integrated into a single behavior tree architecture.

\noindent \textbf{Initialization: }\label{takeoff}At the beginning of each episode, the robot takes off from its starting location and ascends to $5\text{m}$. As in~\cite{yokoyama2024vlfm}, it also performs a $360^\circ$ rotation in its place to obtain a holistic view of the environment. This step ensures a consistent start across trials and sufficient altitude for safe exploration.

\noindent \textbf{Frontier-based Exploration: }\label{frontier-3d} 
The aerial robot defaults to frontier-based exploration when no relevant semantic cues are available. Frontiers $\mathcal{F}_t$ are clustered using DBSCAN with parameters $\epsilon$ and \textit{min\_samples}, yielding a set of frontier centroids $C$. Each centroid is scored by a weighted combination of its distance from the robot’s position and a momentum-based penalty for sharp heading changes. The centroid with the lowest score is then selected as the next waypoint.

\noindent \textbf{Semantic Voxel-based Search: }\label{semantic-voxel-search} As the robot explores, information captured in the RGB images and falls within the depth range is stored in semantic voxels $V_t$. Each voxel $v \in V_t$ contains a semantic feature $\mathbf{f}(v)$ produced by RayFronts encoder. For comparison with target object classes, we adapt these features into the SIGLIP embedding \cite{siglip} space. For a given task, we compute the cosine similarity between SIGLIP-adapted voxel feature $\tilde{\mathbf{f}}(v)$ and the SIGLIP embedding of each target object class query $\mathbf{q}_j$ (e.g., SIGLIP(``water tower")):
\begin{equation}
    s(v,\mathbf{q}_j) = \frac{\tilde{\mathbf{f}}(v) \cdot \mathbf{q}_j}{\|\tilde{\mathbf{f}}(v)\| \, \|\mathbf{q}_j\|}, \quad j=1,\dots,J
\end{equation}

where $J$ is number of object classes in the task. Because multiple text queries can be evaluated in parallel, this formulation naturally supports multi-class navigation. Voxels exceeding a similarity threshold $\epsilon_\text{vox}$ are retained:
\begin{equation}
    V_\text{filtered} = \{v \in V_t \mid \exists~j~~s.t.~s(v,\mathbf{q}_j) > \epsilon_\text{vox} \}
\end{equation}
To approximate object-level regions, connected component labeling (CCL) is applied to $V_\text{filtered}$, grouping adjacent voxels into clusters while filtering out small outliers. 
\begin{equation}
    \{C_1, \dots, C_N\} \leftarrow \textsc{CCL}(V_\text{filtered}) \quad~s.t.~|C_n| \geq \tau_\text{min}
\end{equation}

These clusters serve as candidate object hypotheses and are visited sequentially, starting from the closest. For safe navigation, a global waypoint is set just outside the cluster's bounding box, offset outward by $\Omega$.

\noindent \textbf{Semantic Ray-based Search: }\label{semantic-ray-search} While $V_t$ stores information about nearby objects, outdoor navigation often requires reasoning about objects that lie far beyond the depth range. To capture such long-range cues, we employ semantic rays $R_t$, which encodes information about distant objects that are briefly visible in the RGB images but outside depth coverage. 

A set of rays $R_t=\{r_1,..,r_N\}$ is cast outward in multiple directions beyond the current frontiers $\mathcal{F}_t$. Each ray $r_i$ is associated with semantic features $\mathbf{f}(r_i)$, which is adapted to SIGLIP embeddings $\tilde{\mathbf{f}}(r_i)$. For a given task, we compute cosine similarity between the ray features $\tilde{\mathbf{f}}(r_i)$ and the SIGLIP embedding features of the text queries $\mathbf{q}_j$ corresponding to the target objects:
\begin{equation}
    s(r_i,\textbf{q}_j) = \frac{\tilde{\mathbf{f}}(r_i)\cdot \mathbf{q}_j}{\|\tilde{\mathbf{f}}(r_i)\| \|\mathbf{q}_j\|}
\end{equation}
Rays with similarity above a threshold $\epsilon_{ray}$ form the subset: 
\begin{equation}
    R_\text{filtered} = \{r_i \in R_t | \max_{j} s(r_i,\textbf{q}_j) \geq \epsilon_\text{ray}\}
\end{equation}
 When $R_\text{filtered} \neq \emptyset$,  ray-based search is initiated. The rays in $R_\text{filtered}$ are grouped into angular bins $\{B_1, B_2, ...,B_K\}$. Each group $B_k$ is scored using a weighted combination of proximity $\phi_\text{prox}(B_k)$ and density $\phi_\text{dens}(B_k)$: 
 \begin{equation}
     Score(B_k) = \alpha\cdot \phi_\text{prox}(B_k)  + \beta \cdot \phi_\text{dens}(B_k)
 \end{equation}
 The robot selects the highest-scoring $B_k$ as a global waypoint.

\begin{figure}[t!]
    \centering
\includegraphics[width=1.0\linewidth]{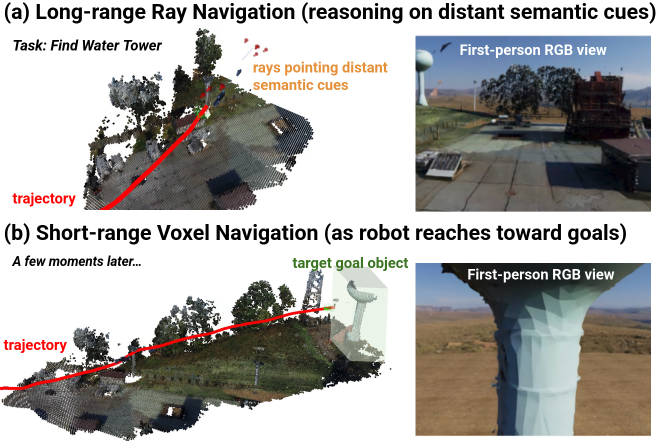}
    \captionsetup{font=small}
    \caption{(a) When no information exists within depth range, the robot activates semantic ray navigation for long-range coarse search, using ray features where the tower was briefly captured in the RGB view. (b) As it follows the rays and approaches the target, voxels naturally form and the search transitions to precise voxel-based navigation.}
    \label{ray-vox-switch}
\end{figure}

As the robot follows selected rays and approaches the object, semantic voxels naturally begin to form, at which point the system transitions smoothly into voxel-based search for more precise localization (Fig.~\ref{ray-vox-switch}). This ray-based search enables long-range reasoning beyond what voxels alone can achieve and provides additional benefit--it introduces a persistent memory: once observed, an object's information remains encoded in the rays even if it is no longer visible in the current image. This motivates our next discussion on memory. 

\noindent \textbf{Task-Agnostic 3D Voxel-Ray  Memory: }\label{memory}
Together, the semantic voxels $V_t$ and semantic rays $R_t$ constitute a unified spatial-semantic memory that accumulates observations over time without overwriting earlier data. This persistent representation explicitly encodes both geometry and semantics, enabling long-horizon planning in large, unstructured environments.

A key feature of our approach is that the voxel-ray memory is task-agnostic, distinguishing it from approaches that rely on task-conditioned 3D maps. In task-conditioned methods, even open-vocabulary models still require a predefined set of tasks (object types or queries) to guide the mapping process. In contrast, our memory is task-agnostic, meaning that no predefined queries are needed during mapping, and the task can even change on the fly. When a new task arises, the robot can align the new query with the existing voxel-ray memory using a similarity score to select relevant voxels and rays.

\begin{figure}[t]
    \centering
\includegraphics[width=1.0\linewidth]{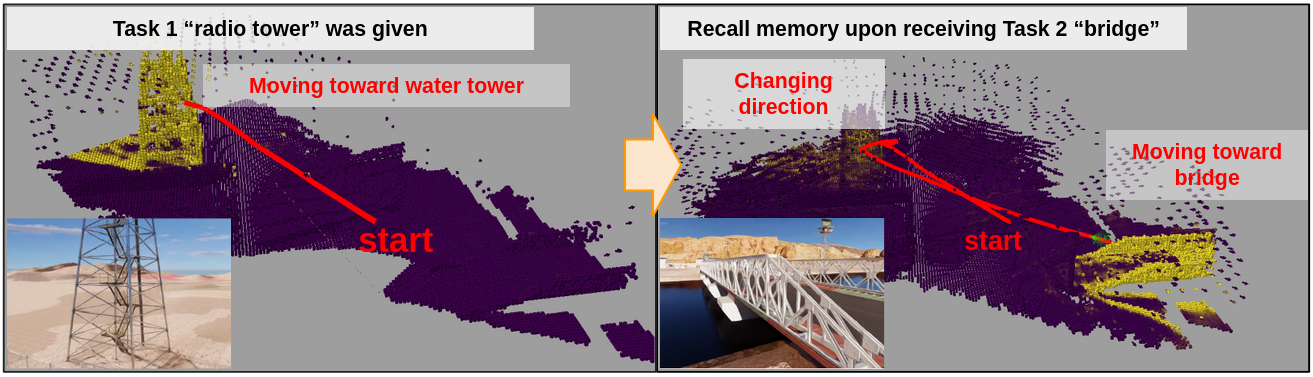}
\captionsetup{font=small}
    \caption{Open-set semantic voxel-ray map serves as memory. When the first task (``radio tower") is given, the robot navigates toward it. Upon introduction of a second task (``bridge"), it queries the existing memory and uses highlighted cues to guide search. Yellow voxels and rays indicate task-relevant highlights.}
    \label{memory}
\end{figure}

This property makes our method particularly suitable for online mapping. As illustrated in Figure~\ref{memory}, while the robot is creating a semantic voxel-ray memory to locate the first task object (e.g., a radio tower), it can seamlessly handle a new task introduced midway (e.g., a bridge) without rebuilding the map. Instead, it simply compares the new query to the existing memory and identifies the most relevant voxels and rays.

\noindent \textbf{LVLM-guided Search: }\label{lvlm-guided-search} While semantic voxels and rays capture information within and beyond the depth sensor range, both can fail if the target objects never appear, even briefly, in any RGB images. To handle such cases, we introduce an additional search strategy guided by an LVLM. At sparse intervals $T_{\mathrm{lvlm}}$, the LVLM is queried with the current image $I_t$ and a prompt $\mathcal{P}$, producing auxiliary objects $aux_j$ that are semantically related to the targets.
\begin{equation}
    \{ aux_j \}_{j=1}^{J_\text{aux}} \leftarrow \textsc{LVLM}(\mathcal{P}, I_t)
\end{equation}
\vspace{-5mm}

These auxiliary cues are converted into additional text queries and fused into the ray-based search, enabling the robot to pursue contextual objects as guidance toward the target. 
\begin{equation}
\begin{aligned}
    Q &= \{\mathbf{q}_j\}_{j=1}^J \cup \{\mathbf{q}_{aux_j}\}_{j=1}^{J_\text{aux}} \\[6pt]
    s(r_i,\textbf{q}_j) &= \frac{\tilde{\mathbf{f}}(r_i) \cdot \mathbf{q}_j}{\|\tilde{\mathbf{f}}(r_i)\|\,\|\mathbf{q}_j\|}, 
    \quad r_i \in R, \; \mathbf{q}_j \in Q
\end{aligned}
\end{equation}
The remaining procedure follows the standard ray-based search strategy. We employ InternVL3-2B \cite{zhu2025internvl3} as the LVLM. 

\noindent \textbf{Behavior Tree Integration: }\label{BT-integration} To coordinate the navigation modules, we use a behavior tree (BT) that structures modular behaviors and ensures robust execution. After initialization, control proceeds through a priority-based sequence:
\begin{itemize}
    \item \textbf{Semantic voxel-based search:} prioritized first, as nearby objects within depth range provide the most reliable cues.
    \item \textbf{Semantic ray-based search:} used when voxel cues are insufficient, using distant observations for long-range coarse guidance.
    \item \textbf{LVLM-guided search:} provides auxiliary cues from sparse LVLM queries when needed.
    \item \textbf{Frontier-based exploration:} fallback strategy to maintain coverage of unexplored regions when semantic cues are unavailable.
\end{itemize}
This modular BT design offers two key benefits: (i) it enables autonomous selection of the most suitable behavior for each state, and (ii) it guarantees resilience and robustness by sequentially activating alternative strategies when one fails.

\subsection{Local Trajectory Planning}
For local trajectory planning and collision avoidance, we use DROAN \cite{dubey2017droan}, a disparity-based algorithm that inflates the configuration space around detected obstacles in the disparity image. The local planner selects the best path from a library of candidate trajectories and, given a global waypoint, generates a collision-free trajectory toward the target. 

\section{SIMULATION EXPERIMENTS}
\begin{figure}[t!]
    \centering
\includegraphics[width=0.85\linewidth]{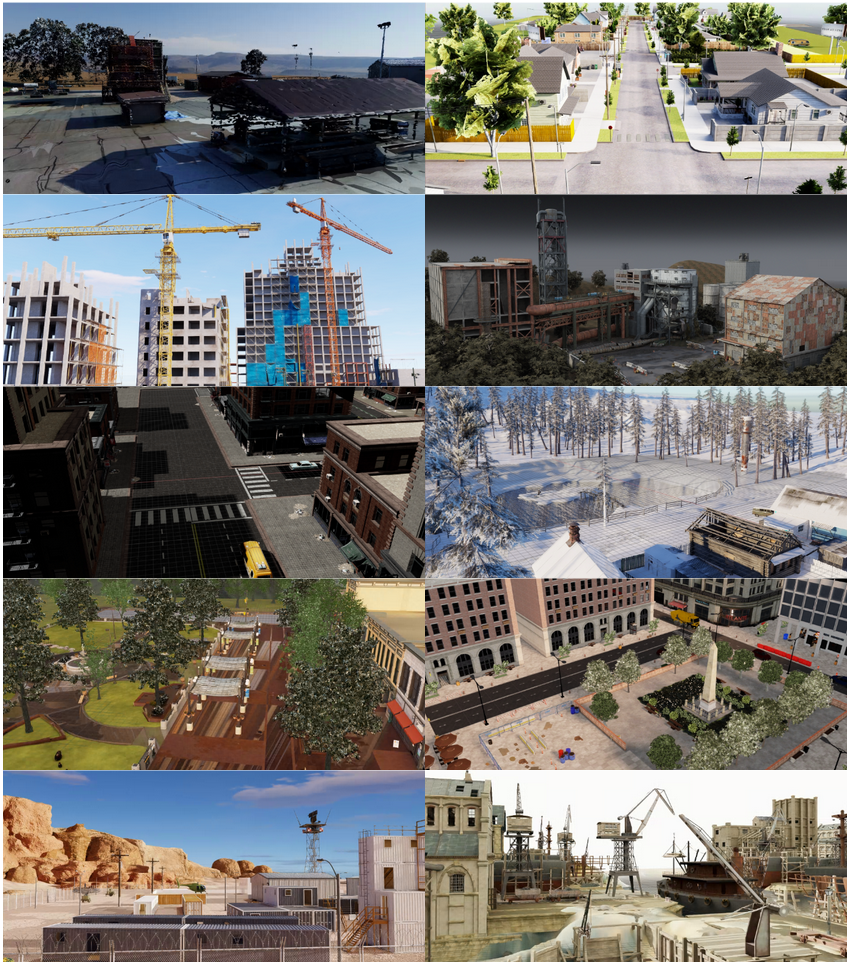}
\vspace{-4pt}
\captionsetup{font=small}
\caption{Ten simulation environments used for evaluation.}
    \label{environments}
\end{figure}
\subsection{Simulator, Environments, and Experiment Setup}
We conduct photorealistic robot simulation using NVIDIA Isaac Sim. Since no standard benchmark exists for outdoor semantic navigation, we design ten environments: one digital twin of a real-world site and nine designed simulated scenes. To introduce a variety of conditons, each environment is tested with three starting poses. The ten simulation environments are shown in Fig.~\ref{environments}. We deploy our custom aerial robot autonomy stack in these environments. The robot model is based on the Spirit platform from Ascent Aerosystems, and all experiments are run on an NVIDIA RTX 6000 Ada GPU.
\begin{figure*}[ht!]
    \centering
\includegraphics[width=1.0\linewidth,trim=0mm 60mm 2mm 0mm]{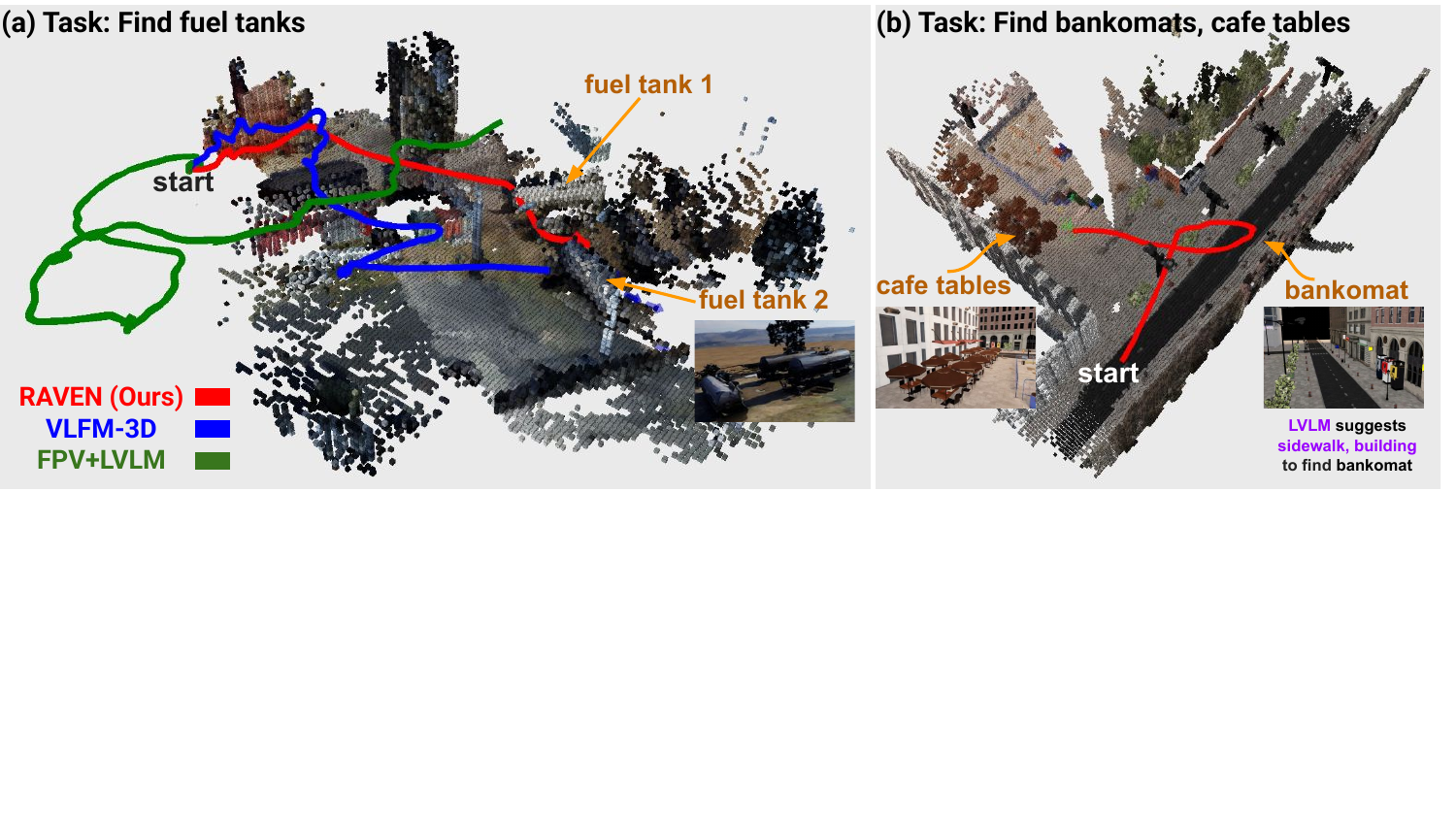}
    \captionsetup{font=small}
    \caption{(a) Trajectories of \MethodName~(red), VLFM-3D (blue), FPV+LVLM (green) for finding fuel tanks. FPV+LVLM relies on recent views and reactive policies, producing a meandering path. VLFM-3D uses a value map but chases momentary maxima, yielding myopic behavior. \MethodName~leverages semantic rays for long-range reasoning, generating an efficient path. (b) \MethodName~trajectory for locating cafe tables and bankomats. Starting with empty memory and failing voxel-ray search, the robot invokes LVLM, which suggests sidewalks and building as auxiliary cues for bankomats. Following rays pointing sidewalk, it finds the bankomats, turns back, spots and navigates to the cafe tables.}
    \label{qualitative-analysis}
\end{figure*}
\subsection{Tasks}
\label{tasks}
Unlike prior works that typically consider single-OGN tasks, our task formulation generalizes to multi-class, multi-instance settings, as well as sequential task scenarios.

\noindent \textbf{Type I: Single-Class Tasks:} All target instances belong to a single class $c \in \mathcal{C}$:
\vspace{-2mm}
\[
\mathcal{G} = \{ g_m \mid g_m \text{ of class } c, \; m=1,\dots,M \}, \quad c \in \mathcal{C}
\]
This includes either a single instance ($M=1$) or multiple instances ($M>1$) of that class.

\noindent \textbf{Type II: Multi-Class Tasks:} Target instances span multiple classes $\mathcal{C}_\text{target} \subseteq \mathcal{C}$:
\vspace{-2mm}
\[
\mathcal{G} = \bigcup_{c \in \mathcal{C}_\text{target}} \{ g_m \mid g_m \text{ of class } c \}
\]
The agent must navigate to all instances in these classes.

\noindent \textbf{Type III: Sequential Dual-Class Tasks:} Target classes are revealed sequentially. Let $\mathcal{G}_1$ and $\mathcal{G}_2$ denote the first and second class instances, respectively. The robot must first find at least one instance of $\mathcal{G}_1$; once found, $\mathcal{G}_2$ is revealed. 
\[
\mathcal{G} = \mathcal{G}_1 \cup \mathcal{G}_2, \quad \text{with } \mathcal{G}_2 \text{ revealed after first $\mathcal{G}_1$ is found}
\]
This task evaluates the robot’s ability to exploit internal maps and past observations to quickly search for the second class.

\subsection{Metrics}
\label{metric}
Single-OGN is commonly evaluted by \textit{Success Rate (SR)}--whether the robot reaches the goal within budget--and its path-efficiency variant \textit{Success weighted by Path Length (SPL)} \cite{chaplot2020object, chang2023goat, yokoyama2024vlfm}. In multi-OGN tasks, however, a binary success criterion fails to capture partial completion. We therefore adopt the \textit{Progress} and \textit{Progress weighted by Path Length (PPL)} metrics, introduced in \cite{wani2020multion}.

\noindent \textbf{Progress:} Let a task specify $M$ target object goals $\mathcal{G} = \{g_1, \dots, g_M\}$ (each instance counts separately). If the robot  reaches $K$ of them within the budget, we define
\[\text{Progress} = \frac{K}{M}.\]
In the single-object case, this reduces to the standard SR.

\noindent \textbf{Progress weighted by Path Length (PPL):} Let $p$ be the length of the executed trajectory up until the $K$-th distinct goal is reached (set PPL$=0$ if $K=0$). PPL normalizes progress by trajectory length $p$ and compares it to the optimal path length $d_K$ visiting the best subset of $K$ goals in the best order. 

\[\text{PPL} \;=\; \frac{d_K}{p} \cdot \frac{K}{M}.\]
In single OGN, PPL reduces to standard SPL.

\subsection{Baselines}
We select three baseline categories for comparison: one representing classical exploration and search, one representing map-free semantic navigation, and one representing map-based semantic navigation as the state-of-the-art baseline. 
\begin{itemize}
    \item Frontier-3D \cite{best2022resilient}: A classical frontier-based exploration method, extended to 3D aerial robot settings. 
    \item FPV+LVLM: A map-free semantic navigation approach that maintains a short history of recent FPV frames and prompts LVLM \cite{zhu2025internvl3} to output discrete action commands.
    \item VLFM-3D \cite{yokoyama2024vlfm}: A 3D extension of the SOTA 2D BEV online OGN method, VLFM. The image encoder and local planning are adopted from our approach. 
\end{itemize}
We also ablate the impact of each component of our method:
\begin{itemize}
    \item Semantic Voxels: Excludes rays and LVLM to assess the effect of ignoring long-range observation encoding
    \item Semantic Rays: Excludes voxels and LVLM to assess the impact of lacking precise in-range localization
    \item Semantic Voxels + Rays: Uses both voxels and rays but no LVLM to measure the benefit of auxiliary cues. 
\end{itemize}

\subsection{Comparison to Baseline Methods}
\begin{table*}[t]
\centering
\caption{Comparison of \MethodName~with baselines. Progress and PPL are reported for each task type.}
\label{tab:baseline-comparisons}
\begin{tabular}{l|cc|cc|cc|cc}
\toprule
 & \multicolumn{2}{c|}{Task Type I (40 tasks)} & \multicolumn{2}{c|}{Task Type II (30 tasks)} & \multicolumn{2}{c|}{Task Type III (30 tasks)} & \multicolumn{2}{c}{All Tasks (100 tasks)} \\
 & Progress(\%) & PPL(\%) & Progress(\%) & PPL(\%) & Progress(\%) & PPL(\%) & Progress(\%) & PPL(\%) \\
\midrule
Frontier-3D \cite{best2022resilient}    & 6.33 & 3.18 & 4.95 & 2.52 & 3.33 & 2.09 & 5.02 & 2.66 \\
FPV+LVLM \cite{zhu2025internvl3} & 17.09 & 7.19 & 11.88 & 5.57 & 8.33 & 3.32 & 12.91 & 5.54 \\
VLFM-3D \cite{yokoyama2024vlfm}       & 35.34 & 19.79 & 27.59 & 15.48 & 24.45 & 10.78 & 29.75 & 15.79 \\
\MethodName& \textbf{54.19} & \textbf{37.28} & \textbf{42.08} & \textbf{25.04} & \textbf{52.78} & \textbf{31.55} & \textbf{50.14} & \textbf{31.89} \\
\bottomrule
\end{tabular}
\end{table*}

\begin{table*}[t]
\centering
\caption{Ablation results of \MethodName. Progress and PPL are reported for each task type.}
\label{tab:ablation}
\begin{tabular}{l|cc|cc|cc|cc}
\toprule
 & \multicolumn{2}{c|}{Task Type I (40 tasks)} & \multicolumn{2}{c|}{Task Type II (30 tasks)} & \multicolumn{2}{c|}{Task Type III (30 tasks)} & \multicolumn{2}{c}{All Tasks (100 tasks)} \\
 & Progress(\%) & PPL(\%) & Progress(\%) & PPL(\%) & Progress(\%) & PPL(\%) & Progress(\%) & PPL(\%) \\
\midrule
Semantic Voxels (No Rays, LVLM)        & 19.26 & 11.92 & 12.52 & 8.21 & 12.22 & 7.64 & 14.63 & 9.52 \\
Semantic Rays (No Voxels, LVLM)         & 39.75 & 28.76 & 30.77 & 17.13 & 41.67 & 20.52 & 34.47 & 22.81 \\
Semantic Voxels + Rays (No LVLM) & 50.05 & 34.55 & 38.87 & 22.99 & \textbf{53.89} & 30.83 & 47.85 & 29.96 \\
\MethodName~(Sem Voxels + Rays + LVLM)  & \textbf{54.19} & \textbf{37.28} & \textbf{42.08} & \textbf{25.04} & 52.78 & \textbf{31.55} & \textbf{50.14} & \textbf{31.89} \\
\bottomrule
\end{tabular}
\end{table*}

\begin{figure*}[ht]
    \centering
\includegraphics[width=0.9\linewidth]{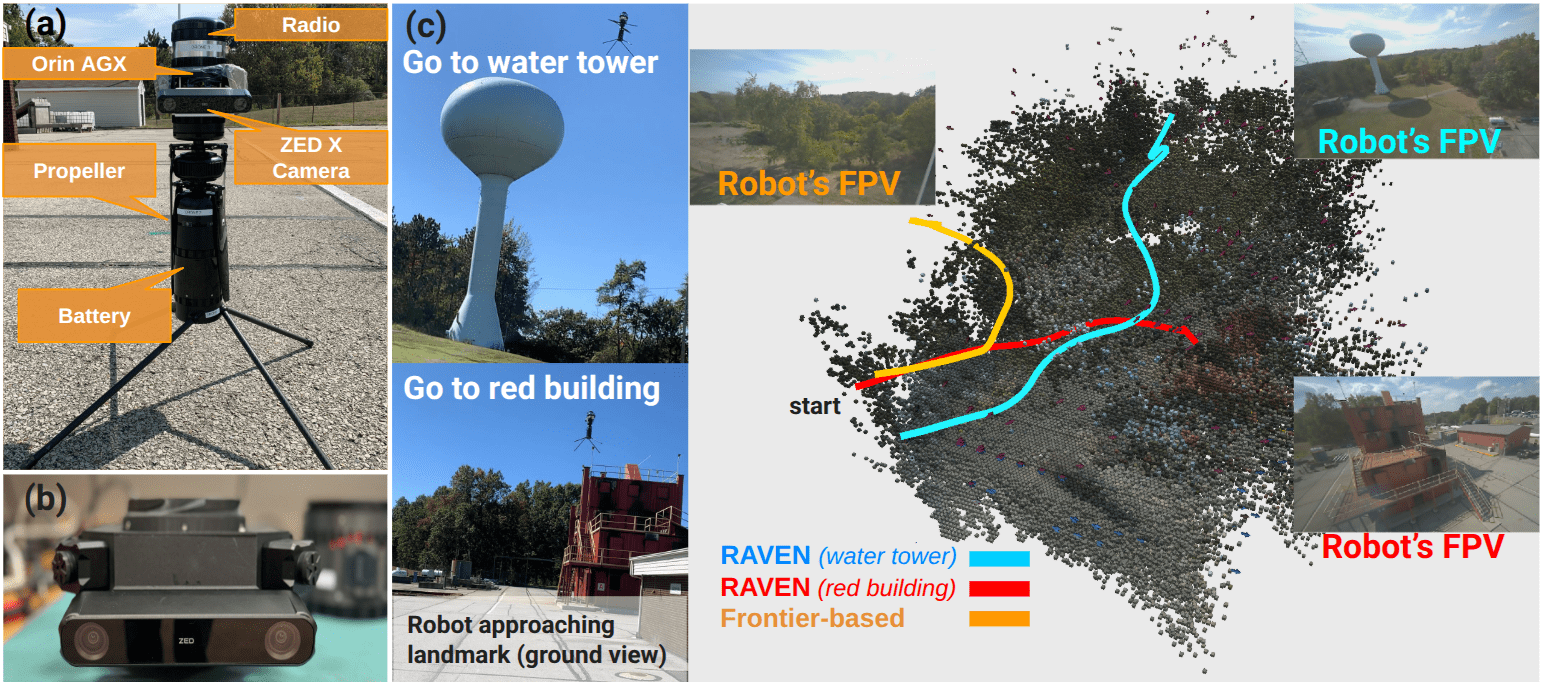} 
\captionsetup{font=small}
\caption{(a) Robot hardware design (b) Front view of compute and camera payloads (c) \MethodName's real-robot navigation trajectories toward the water tower (light blue), red building (red), and the frontier-based exploration trajectory (yellow), and their first-person, ground view images.}
    \label{real-robot}
\end{figure*}

We compare the performance of \MethodName~with all baselines. \MethodName~achieves the best performance across all three task types, followed by the map-based VLFM-3D, then the map-free FPV+LVLM, while the non-semantic frontier-based exploration performs worst, as shown in Table~\ref{tab:baseline-comparisons}. In particular, \MethodName~delivers 68.5\% relative improvement in Progress and 102\% relative improvement in PPL over VLFM-3D. 

Figure~\ref{qualitative-analysis}(a) illustrates single-class navigation to fuel tanks from the same start position with three methods. The map-free FPV+LVLM relies only on recent FPV frames and LVLM prompts; without a consistent map it produces a reactive, meandering path, resulting in low efficiency. VLFM-3D benefits from a value map obtained by projecting similarity scores onto frontiers, which allows eventual goal finding; however, its motion remains myopic, repeatedly chasing the momentary maximum in the value map and yielding an unstable trajectory. In contrast, \MethodName~activates semantic rays as soon as the fuel tank is briefly captured in an RGB image and travels nearly straight to the goal, producing a highly efficient path. 

Figure~\ref{qualitative-analysis}(b) showcases a multi-class task and the role of LVLM. The robot is tasked to find cafe tables and bankomats, neither initially encoded in voxels or rays. Invoking LVLM search suggests additionally seeking sidewalk and building to find the bankomats; incorporating these cues into ray-search quickly guides the robot to the bankomats located on the sidewalk near a building. The robot then reorients and discovers the cafe tables through voxel-ray search. 

\subsection{Ablation Studies}
We perform ablation studies on each component of \MethodName, with results in Table~\ref{tab:ablation}. For Task Types I and II, the full model consistently outperforms all ablated variants, confirming the benefit of every component. For Task Type III, the semantic voxel+ray configuration achieves slightly higher Progress (within the margin of error), indicating that LVLM is less critical when sequential goal changes mainly test memory from voxels and rays. In contrast, LVLM auxiliary cues clearly aid in Task Types I and II.

Overall, semantic rays contribute more strongly than voxels by providing long-range guidance when objects are far away in large scenes, while voxels alone struggle to reason over long distances. Nevertheless, voxel search remains essential for precise localization near a target. Moreover, false positives in ray search can be particularly detrimental, as they may mislead the robot over long distances. These results indicate that voxels and rays are complementary, and both are necessary.

\section{REAL-ROBOT EXPERIMENTS}
\subsection{Hardware Design}
We use an aerial robot based on the Ascent Aerosystems Spirit platform, equipped with custom payloads. These include an NVIDIA Jetson AGX Orin for onboard computation and a ZED X camera providing RGB and depth for online mapping. Due to unreliable onboard pose estimation, we use GPS odometry. The robot components are shown in Figure~\ref{real-robot}(a), and the Orin AGX and ZED X payloads is shown in Figure~\ref{real-robot}(b).

\subsection{Experiments and Results}
We conducted real-world experiments using onboard computing with semantic voxel-ray online mapping and \MethodName, while leaving onboard LVLM integration for future work. Field tests took place at a firefighter training site. \MethodName~was evaluated on two tasks—finding a water tower (beyond depth perception range) and finding a red building—and, for reference, a frontier-based exploration run was also conducted without a specific target. Since no oracle path exists in the real environment, the PPL metric is not applicable; instead, we report qualitative outcomes, trajectory lengths, average flight speed, and real-time mapping rates. The frontier-based method achieved a total flight distance of $40.46\text{m}$ with an average speed of $1.29\text{m/s}$, but wandered without purposeful heading. Using \MethodName, the robot incrementally built a semantic voxel-ray memory and navigated toward the water tower and red building, as shown in Figure~\ref{real-robot}(c). The trajectory lengths for the two tasks were $74.97\text{m}$ and $42.72\text{m}$, respectively, with an average flight speed of $1.21\text{m/s}$. The onboard mapping module operated in real time at an average rate of $4.8\text{Hz}$.

\section{CONCLUSION}
We present \MethodName, a resilient behavior tree framework for aerial semantic navigation in outdoor environments. It uses an open-set semantic voxel-ray memory to enable strategic, long-horizon planning. \MethodName~combines semantic voxel search for precise nearby localization, semantic ray search for long-range reasoning, and LVLM-guided search to mitigate sparse targets. We validate \MethodName~through extensive simulation experiments and demonstrate its real-world applicability with initial aerial robot deployments. 





\section*{ACKNOWLEDGMENTS}
This work was supported by Defense Science and Technology Agency (DSTA) under Contract \#DST000EC124000205. Omar Alama is partially funded by King Abdulaziz
University. We thank Andrew Jong for his dedicated contributions to the development of the core autonomy stack.

\ifdefined\conferenceversion
\else
\appendix
\renewcommand{\thesection}{A\arabic{section}}
\setcounter{table}{0}
\renewcommand{\thetable}{A.\arabic{table}}

\section*{\centering A1. Contribution Statement}

\textbf{Seungchan Kim} led the research and developed the behavior tree architecture and algorithms. Implemented baselines, benchmarks, and evaluation scripts, conducted simulation experiments, ablation studies, and real-robot deployment and field tests, and wrote the majority of the paper.

\textbf{Omar Alama} contributed extensively to the conceptual design of the planner and evaluation framework. Helped optimize and integrate the RayFronts voxel-ray memory, surveyed and helped write related works, and provided feedback on figures and paper writing.

\textbf{Dmytro Kurdydyk} imported scenes into the Isaac Sim simulator, annotated environments for benchmarking, and helped integrate large vision–language models into the system.

\textbf{John Keller} optimized local trajectory planning and obstacle avoidance algorithms, and assisted with real-robot deployment and field tests.

\textbf{Nikhil Keetha} contributed to early discussions, helped brainstorm the pipeline, and assisted with paper writing.

\textbf{Wenshan Wang} provided feedback on the research direction, assisted with benchmarking in Isaac Sim, and offered guidance on writing.

\textbf{Yonatan Bisk} contributed to shaping the early research direction, advised on the use of large vision–language models, and provided writing feedback.

\textbf{Sebastian Scherer} supervised the overall project, guided the research direction, participated in discussions and real-world deployment, and provided feedback on the paper writing, figures, and videos.

\section*{\centering A2. Algorithm Details}
Here we detail the algorithmic procedures of each component of behavior tree architecture.
\begin{algorithm}[h]
\caption{RAVEN Behavior Tree}
\begin{algorithmic}[1]
\Require Time budget $T$, start pose $\mathbf{x}_0$, encoder $E$, task $\mathcal{G}$
\State Take-off at $\mathbf{x}_0$, in-place $360^\circ$ rotation
\State \textbf{for} $t$ in $T$:
\State \hspace{0.5mm} features $\mathbf{f}_t \leftarrow E(I_t)$
\State \hspace{0.5mm} voxels $O_t \leftarrow \textsc{UpdateVoxels}(O_{t-1}, D_t)$
\State \hspace{0.5mm} frontiers  $\mathcal{F}_t \leftarrow \textsc{UpdateFrontiers}(\mathcal{F}_{t-1}, D_t)$
\State \hspace{0.5mm} semantic voxels $V_t \leftarrow \textsc{Project}(\mathbf{f}_t,O_t)$
\State \hspace{0.5mm} semantic rays $R_t \leftarrow \textsc{RayProject}(\mathbf{f}_t,\mathcal{F}_t)$
\State \hspace{0.5mm} \textbf{if} waypoint $\Psi$ is not available:
\State \hspace{1.5mm} \textbf{if} $\textbf{SemanticVoxelSearch}.\textsc{Can\_Execute($V_t, \mathcal{G}$)}:$
\State \hspace{3.0mm} $\Psi \leftarrow \textbf{SemanticVoxelSearch}.\textsc{Execute()}$
\State \hspace{1.5mm} \textbf{else if} $\textbf{SemanticRaySearch}.\textsc{Can\_Execute($R_t, \mathcal{G}$)}:$
\State \hspace{3.0mm} $\Psi \leftarrow \textbf{SemanticRaySearch}.\textsc{Execute()}$
\State \hspace{1.5mm} \textbf{else if} $\textbf{LVLMGuidedSearch}.\textsc{Can\_Execute($R_t, \mathcal{G}$)}:$
\State \hspace{3.0mm} $\Psi \leftarrow \textbf{LVLMGuidedSearch}.\textsc{Execute()}$
\State \hspace{1.5mm} \textbf{else:} $\Psi \leftarrow  \textbf{FrontierBasedExploration}.\textsc{Execute($\mathcal{F}_t$)}$
\State \hspace{0.5mm} $a_t \leftarrow \text{local planner}(\Psi)$, transition to $\mathbf{x}_{t+1}$
\Ensure Trajectory ($\mathbf{x}_0, \mathbf{x}_1, ..., \mathbf{x}_t$), where $t \leq T$
\end{algorithmic}
\end{algorithm}

\begin{algorithm}[h]
\caption{Semantic Voxel Search}
\begin{algorithmic}[0]
\Require $V_t$, $\mathcal{G}$
\State \textbf{def} Can\_Execute($V_t, \mathcal{G}$):
\State \hspace{1.5mm} $c_1,..,c_J \leftarrow \mathcal{G}$ \Comment{classes}
\State \hspace{1.5mm} $\mathbf{q}_j \leftarrow \textsc{SIGLIP}(c_j) \quad \forall j$ \Comment{queries features}
\State \hspace{1.5mm} align $\tilde{\mathbf{f}}(v) \leftarrow \mathbf{f}(v) \quad \forall v \in V_t$
\State \hspace{1.5mm} $s(v,\mathbf{q}_j)=\tilde{\mathbf{f}}(v) \cdot \mathbf{q}_j / \| \tilde{\mathbf{f}}(v)\| \| \mathbf{q}_j\|$ \Comment{similarity scoring}
\State \hspace{1.5mm} $V_\text{filtered}=\{v \in V_t \mid \exists~j~~s.t.~s(v,\mathbf{q}_j) > \epsilon_\text{vox} \} $ \Comment{filtering}
\State \hspace{1.5mm} $\{C_1,..,C_N\} \leftarrow \textsc{CCL}(V_\text{filtered}) \quad~s.t.~|C_n| \geq \tau_\text{min}$
\State \hspace{1.5mm} $\{U_k\} \leftarrow \textsc{Unvisited}(\{C_1,..,C_N\})$
\State \hspace{1.5mm} \textbf{if} $|\{U_k\}| \geq 1$: \textbf{return} True \textbf{else}: \textbf{return} False
\State \textbf{def} Execute():
\State \hspace{1.5mm}$U^\ast \leftarrow \arg\min_{U_k} \textsc{dist}(U_k.\mathrm{center}, \mathrm{cur\_pose})$
\State \hspace{1.5mm}$\mathrm{dir} \leftarrow U^\ast.\mathrm{center} - \mathrm{cur\_pose}$,
      $\mathrm{dir\_norm} \leftarrow \mathrm{dir}/\|\mathrm{dir}\|$
\State \hspace{1.5mm}$p_\text{surf} \leftarrow \textsc{RayBoxIntersect}(\mathrm{cur\_pose}, U^\ast.\mathrm{center}, U^\ast.\mathrm{size})$
\State \hspace{1.5mm}$p_\text{adj} \leftarrow p_\text{surf} - \Omega\cdot\mathrm{dir\_norm}$
\State \hspace{0.5mm} \textbf{return} $\Psi \leftarrow p_\text{adj}$  \Comment{Waypoint}
\end{algorithmic}
\end{algorithm}

\begin{algorithm}[h]
\caption{Semantic Ray Search}
\begin{algorithmic}[0]
\Require $R_t$, $\mathcal{G}$
\State \textbf{def} Can\_Execute($R_t, \mathcal{G}$):
\State \hspace{1.5mm} $c_1,..,c_J \leftarrow \mathcal{G}$ \Comment{classes}
\State \hspace{1.5mm} $\mathbf{q}_j \leftarrow \textsc{SIGLIP}(c_j) \quad \forall j$ \Comment{queries features}
\State \hspace{1.5mm} align $\tilde{\mathbf{f}}(r_i) \leftarrow \mathbf{f}(r_i) \quad \forall r_i \in R_t$
\State \hspace{1.5mm} $s(r_i,\mathbf{q}_j)=\tilde{\mathbf{f}}(r_i) \cdot \mathbf{q}_j / \| \tilde{\mathbf{f}}(r_i)\| \| \mathbf{q}_j\|$ \Comment{similarity scoring}
\State \hspace{1.5mm} $R_\text{filtered}=\{r_i \in R_t | \max_{j} s(r_i,\textbf{q}_j) \geq \epsilon_\text{ray}\} $ \Comment{filtering}
\State \hspace{1.5mm} \textbf{if} $R_\text{filtered} \neq \emptyset$: \textbf{return} True \textbf{else}: \textbf{return} False
\State \textbf{def} Execute():
\State \hspace{1.5mm} $\mathrm{cur.xy} \leftarrow$ XY components of $\mathrm{cur\_pose}$
\State \hspace{1.5mm} $R_{\text{valid}} \leftarrow \{r_i \in R_{\text{filtered}} \mid r_i.\mathrm{dir} \cdot (r_i.\mathrm{orig} + r_i.\mathrm{dir} - \mathrm{cur.xy}) > 0 \}$
\State \hspace{1.5mm} bins $B = [\hspace{1.0mm}]$
\State \hspace{1.5mm} \textbf{for} each $r_i \in R_{\text{valid}}$:
\State \hspace{3.0mm} $r_i.\mathrm{assigned}=\text{False}$
\State \hspace{4.5mm} \textbf{for} bin $B_k$ in $B$:
\State \hspace{6.0mm} \textbf{if} $B_k.\mathrm{centroid} \cdot r_i.\mathrm{dir} \geq \theta_\mathrm{thresh}$:
\State \hspace{7.5mm} $B_k.\mathrm{rays}.\text{append}(r_i)$, $B_k.\mathrm{centroid}=\text{mean}(B_k.\mathrm{rays})$
\State \hspace{7.5mm} $r_i.\mathrm{assigned}=\text{True}$
\State \hspace{6.0mm} \textbf{if} not $r_i.\mathrm{assigned}$:
\State \hspace{7.5mm} $B.\text{append}(\{\mathrm{centroid}:r_i, \mathrm{rays}:[r_i]\})$
\State \hspace{1.5mm} $B^\ast \leftarrow \arg\max_{B_k \in B} \; \alpha \cdot \phi_\text{prox}(B_k) + \beta \cdot \phi_\text{dens}(B_k)$
\State \hspace{0.5mm} \textbf{return} $\Psi \leftarrow B^\ast.\mathrm{rays.orig} + B^\ast.\mathrm{rays.dir} \cdot \Omega_\mathrm{ray}$  \Comment{Waypoint}
\end{algorithmic}
\end{algorithm}

\begin{algorithm}[h!]
\caption{LVLM-Guided Search}
\begin{algorithmic}[0]
\Require $R_t$, $\mathcal{G}$
\State \textbf{def} Can\_Execute($R_t, \mathcal{G}$):
\State \hspace{1.5mm} Prompt $\mathcal{P} \leftarrow \mathcal{G}$
\State \hspace{1.5mm} $\{ aux_j \}_{j=1}^{J_\text{aux}} \leftarrow \textsc{LVLM}(\mathcal{P}, I_t)$ \Comment{Auxiliary Objects}
\State \hspace{1.5mm} $\mathcal{G}_\text{aug} \leftarrow \mathcal{G} \cup \{ aux_j \}_{j=1}^{J_\text{aux}}$ \Comment{Augmented Task}
\State \hspace{1.5mm} \textbf{return} $\textbf{SemanticRaySearch}.\textsc{Can\_Execute($R_t, \mathcal{G}_\text{aug}$)}$
\State \textbf{def} Execute():
\State \hspace{1.5mm} \textbf{return} $\textbf{SemanticRaySearch}.\textsc{Execute()}$
\end{algorithmic}
\end{algorithm}

\begin{algorithm}[h]
\caption{Frontier-based Exploration}
\begin{algorithmic}[0]
\Require $\mathcal{F}_t$
\State \textbf{def} Execute():
\State \hspace{1.5mm} $\mathcal{F}_\mathrm{filtered} \leftarrow \{f \in \mathcal{F}_t \ | f.z > z_\text{thresh}\}$
\State \hspace{1.5mm} $C \leftarrow \textsc{DBSCAN}(\mathcal{F}_t, \epsilon, min\_samples)$ \Comment{frontier centroids}
\State \hspace{1.5mm} \textbf{for} $c_f$ in $C$:
\State \hspace{3.0mm} $\mathrm{front\_head} \leftarrow (c_f-\mathrm{cur\_pose})/\|(c_f-\mathrm{cur\_pose})\|$
\State \hspace{3.0mm} $c_f.\mathrm{head} \leftarrow 1.0 - \mathrm{cur\_head} \cdot \mathrm{front\_head}$
\State \hspace{3.0mm} $c_f.\mathrm{score} \leftarrow \alpha_\mathrm{dist} \cdot \textsc{dist}(c_f, \mathrm{cur\_pose}) + \alpha_\mathrm{head} \cdot c_f.\mathrm{head}$
\State \hspace{1.5mm} \textbf{return} $\Psi \leftarrow \arg \min_{c_f} c_f.\mathrm{score}$
\end{algorithmic}
\end{algorithm}

\section*{\centering A3. LVLM Prompt}
\section{LVLM Prompt}
The exact prompt used for the LVLM is shown below:

\begin{verbatim}
prompt = (
 f'<image>\nFind {self._target_objects}.'
 f'List three unique objects or areas 
 that are most helpful as clues or context
 to locate the {self._target_objects}. '
 f'Write ONLY the object or area names
 as a plain comma-separated list.')
\end{verbatim}

\section*{\centering A4. Hyperparameters for Experiments}
Here we list the hyperparameters used for both the simulation and real-robot experiments. For the encoder and mapper of RayFronts, any hyperparameters not specified here follow the default settings of \cite{alama2025rayfronts}. Likewise, any hyperparameters not detailed for the real-robot experiments are identical to those used in the simulation.
\begin{table}[h!]
\caption{Simulation Experiment Hyperparameters}\label{tab:hyperparameters}
\centering
\begin{tabular}{ll}
\hline
\multicolumn{1}{c}{parameter}              & value                  \\ \hline
$\text{Image(RGB,Depth) Resolution}$       & 448x448                \\
$\text{Maximum Depth Range}$               & 30m                    \\
$\text{Frame skip}$                        & 10                     \\
$\text{Frontier DBSCAN}\quad \epsilon$          & 2.7                    \\
$\text{Frontier DBSCAN}\quad min\_samples$      & 3.0                    \\
$\text{Frontier vertical threshold}\quad z_\mathrm{thresh}$ & 1.5m        \\
$\text{Frontier distance weight}\quad \alpha_\mathrm{dist}$ & 1.0        \\
$\text{Frontier heading change weight}\quad \alpha_\mathrm{head}$ & 5.0        \\
$\text{Voxel size}$ & 0.5m \\
$\text{Voxel filtering threshold}\quad \epsilon_\text{vox}$ & 0.98 \\
$\text{Voxel cluster outlier threshold}\quad \tau_\text{min}$ & 30 \\
$\text{Voxel cluster bounding box outward offset}\quad \Omega$ & 1.0\\
$\text{Ray filtering threshold}\quad \epsilon_\text{ray}$ & 0.95 \\
$\text{Ray binning angle threshold}\quad \theta_\text{thresh}$ & $45^\circ$ \\
$\text{Ray proximity parameter}\quad \alpha$ & 1.0 \\
$\text{Ray density parameter}\quad \beta$ & 5.0 \\
$\text{Ray waypoint parameter}\quad \Omega_\mathrm{ray}$ & 6.0 \\
$\text{LVLM invoke period}\quad T_{\mathrm{lvlm}}$ & 20s \\
$\text{LVLM number of auxiliary objects}\quad J_\mathrm{aux}$ & 3 \\
\end{tabular}
\end{table}

\begin{table}[h]
\caption{Real-Robot Experiment Hyperparameters}\label{tab:real-hyperparameters}
\centering
\begin{tabular}{ll}
\hline
\multicolumn{1}{c}{parameter}              & value                  \\ \hline
$\text{Maximum Depth Range}$               & 20m                    \\
$\text{Frontier vertical threshold}\quad z_\mathrm{thresh}$ & 4m        \\
$\text{Ray filtering threshold}\quad \epsilon_\text{ray}$ & 0.7 \\
$\text{Ray waypoint parameter}\quad \Omega_\mathrm{ray}$ & 8.0 \\
\end{tabular}
\end{table}

\section*{\centering A5. Simulation Environments and Tasks Detail}
Here we describe the tasks for each simulation environment. \textbf{Task I} is a single-class object-goal navigation task, \textbf{Task II} is a multi-class object-goal navigation task, and \textbf{Task III} is a sequential dual-class task-switching navigation task.
\begin{table}[h]
\caption{Environments and Tasks}\label{tab:env-tasks}
\centering
\begin{tabular}{l l}
\hline
\multicolumn{1}{c}{Environment} & \multicolumn{1}{c}{Tasks} \\ \hline
Fire Academy & 
\makecell[l]{\textbf{Task I}: ``water tower'', ``radio tower'',\\ ``green container'', ``fuel tank'',\\
\textbf{Task II}: ``water tower, fuel tank'',\\ ``radio tower, blue container'', \\``red container, water tower'' \\
\textbf{Task III}: ``water tower$\rightarrow$radio tower'',\\ ``radio tower$\rightarrow$fuel tank'', ``green container$\rightarrow$water tower''}\\
\hline
Neighborhood & 
\makecell[l]{\textbf{Task I}: ``blue container'', ``bus stop'', \\``yellow car'', ``house'',\\
\textbf{Task II}: ``bus stop, blue container'', ``red container, \\ yellow car, billboard'', ``tunnel, blue container, red car'' \\
\textbf{Task III}: ``bus stop$\rightarrow$blue container'',\\ ``yellow car$\rightarrow$billboard'', ``blue container$\rightarrow$tunnel''}\\
\hline
\makecell[l]{Construction\\Site} & 
\makecell[l]{\textbf{Task I}: ``blue tarp'', ``orange towercrane'', \\``cabling winches'', ``bio toilet'',\\
\textbf{Task II}: ``orange towercrane, forklift'', \\``construction lift, asphalt roller'',\\ ``blue tarp, bio toilet, yellow towercrane'' \\
\textbf{Task III}: ``blue tarp$\rightarrow$orange towercrane'', ``orange\\ towercrane$\rightarrow$blue tarp'', ``cabling winches$\rightarrow$bio toilet''}\\
\hline
\makecell[l]{Abandoned\\Factory} & 
\makecell[l]{\textbf{Task I}: ``white silo'', ``water tower'', ``pipe'',``building"\\
\textbf{Task II}: ``water tower, pipe'', ``white silo, building'',\\ ``building, pipe'' \\
\textbf{Task III}: ``water tower$\rightarrow$white silo'', ``building\\$\rightarrow$white silo'', ``white silo$\rightarrow$pipe''}\\
\hline
\makecell[l]{Abandoned\\City} & 
\makecell[l]{\textbf{Task I}: ``building'', ``car'', ``bus stop'',\\``yellow motorhome"\\
\textbf{Task II}: ``yellow motorhome, car'', ``car, bus stop'',\\ ``bus stop, yellow motorhome'' \\
\textbf{Task III}: ``building$\rightarrow$car'', ``car$\rightarrow$bus stop'', \\``bus stop$\rightarrow$car''}\\
\hline
\makecell[l]{Snowy\\Village} & 
\makecell[l]{\textbf{Task I}: ``human'', ``bridge'', ``pond'',``car"\\
\textbf{Task II}: ``tower, car'', ``human, outhouse'', \\``outhouse, pond'' \\
\textbf{Task III}: ``pond$\rightarrow$tower'', ``tower$\rightarrow$bridge",\\ ``outhouse$\rightarrow$car''}\\
\hline
\makecell[l]{Downtown\\West} & 
\makecell[l]{\textbf{Task I}: ``fountain'', ``fire hydrant'', ``food cart'',\\``trash bin"\\
\textbf{Task II}: ``food cart, trash bin'', ``fountain, food cart'', \\``fountain, trash bin'' \\
\textbf{Task III}: ``fountain$\rightarrow$food cart'', ``trash bin$\rightarrow$fountain",\\ ``food cart$\rightarrow$fountain''}\\
\hline
\makecell[l]{Modern City\\Downtown} & 
\makecell[l]{\textbf{Task I}: ``obelisk'', ``yellow truck'', ``cafe table'',\\``bankomat"\\
\textbf{Task II}: ``obelisk, yellow truck'', ``cafe table,\\ bankomat'', ``bankomat, yellow truck'' \\
\textbf{Task III}: ``cafe table$\rightarrow$obelisk'', ``obelisk$\rightarrow$bankomat",\\ ``yellow truck$\rightarrow$cafe table''}\\
\hline
\makecell[l]{Military\\Base} & 
\makecell[l]{\textbf{Task I}: ``radio tower'', ``helicopter'', ``ATV'',``bridge"\\
\textbf{Task II}: ``radio tower, guard tower'', \\``helicopter, radio tower, mobile radar'',\\ ``radio tower, bridge, helicopter'' \\
\textbf{Task III}: ``guard tower$\rightarrow$helicopter'', ``radio tower\\$\rightarrow$bridge",``bridge$\rightarrow$guard tower''}\\
\hline
\makecell[l]{Shipyard} & 
\makecell[l]{\textbf{Task I}: ``ship'', ``crane'', ``silo'',``ship construction"\\
\textbf{Task II}: ``ship, crane'', ``ship silo'',\\ ``ship construction, crane'' \\
\textbf{Task III}: ``crane$\rightarrow$ship'', ``ship$\rightarrow$silo",``silo$\rightarrow$ship''}\\
\hline
\end{tabular}
\end{table}

\section*{\centering A6. Implementation Details on Baselines}
\textbf{VLFM-3D}: Since the original VLFM was proposed only in a 2D version in prior work~\cite{yokoyama2024vlfm}, we extended it to a 3D version to fit our setting. Fortunately, we already have a 3D frontier set $\mathcal{F}_t$. We projected the encoder features onto each frontier point to construct a 3D cosine-similarity value map. As mentioned earlier, the image encoder and the local trajectory planning for obstacle avoidance were implemented using our own method, so that we could isolate and test the efficacy of the 3D value map component only. The robot moves toward the frontier point with the highest score in the value map.

\textbf{FPV+LVLM}: As mentioned earlier, we used InternVL3-2B~\cite{zhu2025internvl3} as the backbone. We stored the most recent 5 image frames in an image-frame memory and the most recent 200 timesteps of poses in a pose memory. Based on the recent first-person-view images, we prompted the LVLM to choose one of three actions: move forward, turn left, or turn right. Once an action was selected, local planning to reach the corresponding waypoint was carried out using our method.

\section*{\centering A7. Ablation Figures: Voxel vs. Ray Search}
Here we additionally report two figures comparing the trajectories of the ablated components: semantic voxel search only and semantic ray search only. Figures \ref{voxray-compare1} and \ref{voxray-compare2} visualize the differences between the two methods in scenarios where the robot navigate to houses in the Neighborhood and buildings in the Abandoned City environments, respectively (semantic voxel search only: yellow, semantic ray search only: red).

As shown in these figures, the voxel-based search performs short-range, reliable navigation by visiting the nearest detected objects one by one. It provides precise localization but is less effective at covering long distances within the given time. In contrast, the ray-based search tends to pursue faraway objects within its long-range field of view, producing generally straighter trajectories, but it often skips nearby targets. This highlights the need for both components in \MethodName.

\begin{figure}[H]
    \centering
\includegraphics[width=0.95\linewidth]{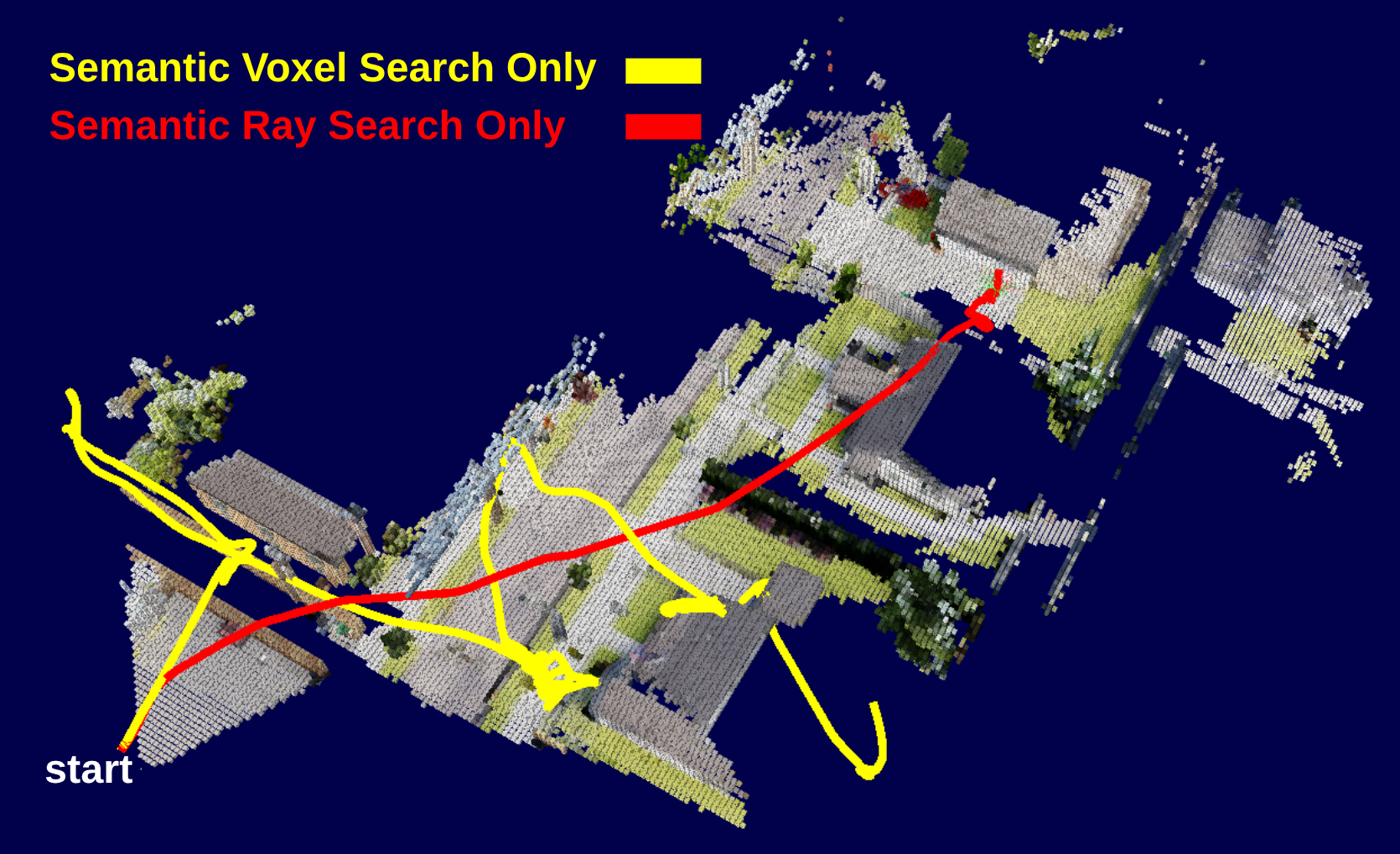}
\captionsetup{font=small}
    \caption{Comparison of semantic voxel search and ray search in the Neighborhood environment, where the goal is to navigate to houses.}
    \label{voxray-compare1}
\end{figure}

\begin{figure}[H]
    \centering
\includegraphics[width=0.95\linewidth]{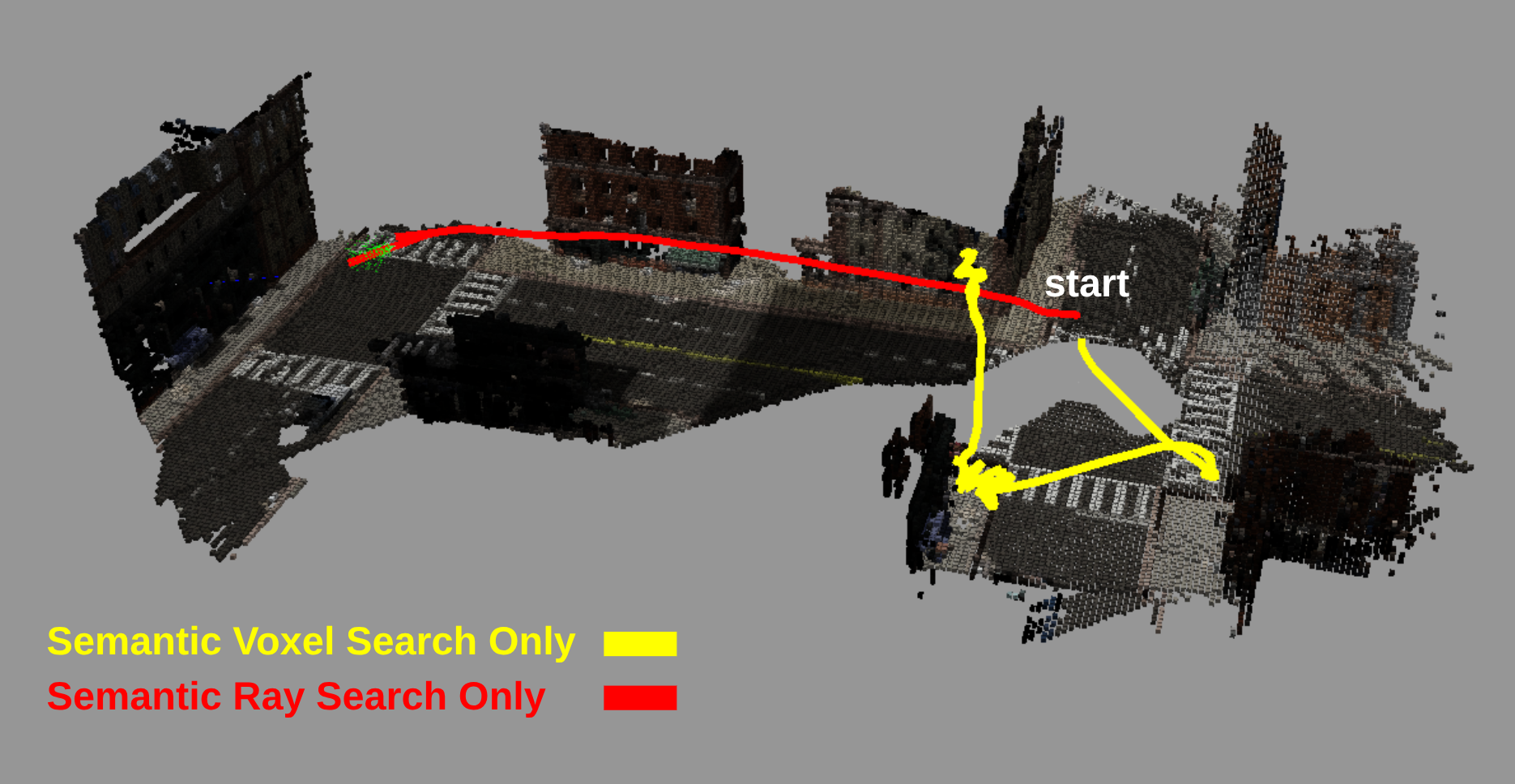}
\captionsetup{font=small}
    \caption{Comparison of semantic voxel search and ray search in the Abandoned City environment, where the goal is to navigate to buildings.}
    \label{voxray-compare2}
\end{figure}

\fi

\footnotesize{
\bibliographystyle{IEEEtran}
\bibliography{IEEEabrv, IEEEexample}
}

\end{document}

%% file: includes/packages.tex
\usepackage{graphics} 
\usepackage{epsfig} 
\usepackage{mathptmx} 
\usepackage{times} 
\usepackage{amsmath} 
\usepackage{amssymb}  
\usepackage[
colorlinks=true,
linkcolor=blue,
citecolor=blue,
filecolor=red,
urlcolor=blue]{hyperref}
\usepackage{algorithm}
\usepackage{eucal}
\usepackage[usenames,dvipsnames]{xcolor}
\usepackage{algpseudocode}
\usepackage{verbatim}
\usepackage{tabularx}
\usepackage{mathrsfs}
\usepackage{caption}
\usepackage{balance}
\usepackage{booktabs} 
\usepackage{tabularx}
\usepackage[numbers,sort&compress]{natbib}
\usepackage{subcaption}
\usepackage{makecell}
\usepackage{float}

%% file: includes/macros.tex
\newcommand{\xxnote}[3]{}
\ifx\hidenotes\undefined
  \renewcommand{\xxnote}[3]{\color{#2}{#1: #3}}
\fi